\documentclass[a4]{article}
\usepackage{authblk}
\usepackage[utf8]{inputenc}
\usepackage[T1]{fontenc}
\usepackage{graphicx}
\usepackage{float}
\usepackage{listings}
\usepackage{multirow}
\usepackage{cleveref}
\usepackage{todonotes}
\usepackage{graphicx}
\usepackage[hang,small,bf,tight]{subfigure}

\begin{document}

\title{Image analysis for automatic measurement of crustose lichens}


\author[1]{Pedro Guedes}
\author[2]{Maria Alexandra Oliveira}
\author[2]{Cristina Branquinho}
\author[1]{João Nuno Silva}

\affil[1]{INESC-ID, Instituto Superior Técnico, Universidade de Lisboa, 1000-029 Lisboa, Portugal}
\affil[2]{Centre for Ecology, Evolution and Environmental Changes, Faculdade de Ciências da Universidade de Lisboa, 1749-016 Lisboa, Portugal}
\affil[ ]{\textit {	pedro.p.guedes@tecnico.ulisboa.pt , 
 \{maoliveira, cmbranquinho\}@fc.ul.pt,  joao.n.silva@inesc-id.pt}}

\maketitle

\begin{abstract}
Lichens, organisms resulting from a symbiosis between a fungus and an algae, are frequently used as age estimators, especially in recent geological deposits and archaeological structures, using the correlation between lichen size and age. Current non-automated manual lichen and measurement (with ruler, calipers or using digital image processing tools) is a time-consuming and laborious process, especially when the number of samples is high.


This work presents a workflow and set of image acquisition and processing tools developed to efficiently identify lichen thalli in flat rocky surfaces, and to produce relevant lichen size statistics (percentage cover, number of thalli, their area and perimeter).  
 
The developed workflow uses a regular digital camera for image capture along with specially designed targets to allow for automatic image correction and scale assignment. After this step, lichen identification is done in a flow comprising assisted image segmentation and classification based on interactive foreground extraction tool (GrabCut) and automatic classification of images using Simple Linear Iterative Clustering (SLIC) for image segmentation and Support Vector Machines (SV) and Random Forest classifiers.  
 
Initial evaluation shows promising results. The manual classification of images (for training) using GrabCut show an average speedup of 4 if compared with currently used techniques and presents an average precision of 95\%. The automatic classification using SLIC and SVM with default parameters produces results with average precision higher than 70\%.  
The developed system is flexible and allows a considerable reduction of processing time, the workflow allows it applicability to data sets of new lichen populations. 
\end{abstract}


\section{Introduction}
\label{chap:intro}

The classification of saxicolous lichen thalli is extremely relevant for dating applications, to estimate the age of surface exposure using a technique named lichenometry. This technique has been mostly applied to rocky surfaces and has been widely used in the study of recent geological, especially glacial, and periglacial deposits \cite{palacios2021,birkeland1982subdivision,carrara1975holocene,garibotti2009lichenometric,hansen2008application,o2003rhizocarpon,Orwin_et_al_2008,pendleton2017using,proctor1983sizes,roberts2010establishing,roof2011indirect,rosenwinkel2015limits,trenbirth2010lichen}. It is based on the relationship between the size of lichens and their age (the largest the lichen, the older it is, and its age can be used to infer the age of exposure of the rock colonized rock surface).

Lichenometry has been essentially based on lichens with circular growth, such as the genus \textit{Rhizocarpon} \cite{doi:10.1177/030913338500900202}.  
This technique is cost-effective and has the added advantage of allowing the estimation of the age of a large set of samples, thus providing statistical robustness. In fact, alternative surface exposure dating techniques applied to rock surfaces, such as cosmogenic isotopes or optically stimulated luminescence, are time and resource consuming and expensive, especially if the number samples is large \cite{ivy2008surface,gliganic2019osl}. 

However, currently the application of lichenometry is not without difficulties. 
Traditional data collection includes lichen thalli selection based on visual inspection and manual measurement of the lichen thalli diameters using a ruler or caliper, on site \cite{doi:10.1177/030913338500900202}. 
This encompasses a series of subjective processes which vary for different people, including the number of thalli to consider (e.g., five largest thalli colonizing a surface), the choice of lichen size parameter (e.g., largest inscribed circle or the largest axis\cite{article}), and, given that this assessment is based in visual inspection, the placement of the measured line. 
As a result, inconsistencies have been found in lichen growth rates derived by different authors, in part due to high inter-operator variance within the data collecting processes. 
In addition, the lack of systematic visual records of observations also lead to low reproducibility. 
The combination of these factors has contributed to a loss of confidence in the technique, which has been branded as pseudo-science \cite{osborn2015lichenometric}. 

Building a robust lichen growth model requires measuring and evaluating its consistency in space, since the growth rate of lichens depends on several factors, such as species and climate, just to list a few \cite{matthews2011, 85f9b2036d5d48ac8a21b8e7c275886e}. 
The growth curves representative of a region are obtained based on the diameter of lichens with known ages. Sampling for application in lichenometry requires identification of lichen species as well as estimation of the thallus, to characterize the population on the surface of interest.

Recently, field data collection and lichen thalli measurements have been made using digital photographs\cite{NicoleM.Henry:2011, TheresaA.Bukovics:2016, MCCARTHY2021107736, maoliveira:2020}, contributing to minimize sources of error related with inter-operator variance and increasing reproducibility. In fact, technological advances in recent decades, such as CMOS image sensors, allow easy access to high definition and low-cost digital cameras that are nowadays part of every field-kit, and an essential tool for recording observations.

In what concerns lichenometry, digital cameras together with  image editing software have been used to measure individual thalli axes, perimeter, and to obtain thallus area \cite{NicoleM.Henry:2011}.
The use of color selection tools, together with geographic information systems, allows to vectorize and scale regions of interest in photographs, in this case lichen thalli, and to convert pixels into areas of lichen thalli \cite{doi:10.1080/15230430.2001.12003411, maoliveira:2020}.




Although most studies use lichen diameter as size parameters, the use of area has been recently indicated as preferable two-dimensional measurement of lichen size and growth, as it includes growth around the entire perimeter of the thallus \cite{maoliveira:2020, roof2011indirect}.  
In fact, crustose lichens develop perfectly attached to the substrate, and they grow essentially spread (flattened shape) along the boundaries of the thalli and marginally in thickness \cite{hill_1981,Seminara}. 
In addition, most crustose lichen species are not perfectly circular, with irregular but significant changes in lichen growth occurring along the perimeter of the thallus \cite{matthews2011}. 
These changes cannot be fully incorporated in measurements when measuring lichen size using diameter instead of area \cite{roof2011indirect}.

Image classification using classification algorithms and learning-based algorithms have been implemented for a wide range of applications and scales.
From the landscape scale, such as classification of land use and plant ecological units (co-occurring plant species) based on multi-spectral satellite imagery \cite{aghababaei2021classification, hurskainen2019auxiliary, akar2012classification}, to the hand specimen scale, such as medical image detection and diagnosis of radiology data \cite{wang2012machine}, and reaching the microscopical scale, such as cell image classification for medical diagnose \cite{gupta2018feature}. 

Several computer vision techniques have been applied in the context of vegetation analysis. These techniques generally involve image segmentation and classification with the interest of measuring certain geometric features such as leaf area, height or volume (to estimate growth), as well as identifying species or possible plants affected by disease \cite{6910608, plant_diseases}.

In one of such works the Otsu's method\cite{4310076} is used to automate lettuce area measurements \cite{LinKaiyan:2014}. 
The proposed work performs a binary segmentation to identify the background and lettuce leaves. 
Otsu's method is used to automatically select the threshold of a greyscale image and it returns a single threshold intensity that separates the pixels into two classes foreground (lettuce leaves) and the background. 
Due to the characteristics of lichen and the rocky surfaces where it grows this technique is not effective. 
One other work directly applied to lichen~\cite{salehi:2016} defines of a vegetation index for lichens based on hyperspectral measurements in the visible to mid-infrared spectrum using samples as training and validation data sets to find the optimal values by minimizing the {RMSE}. 
By using bands narrower that those of RGB digital cameras and in the infrared spectrum it is possible to distinguish lichen from the background. 

Although the use of computer vision techniques has been applied in the solution of similar problems (identification and individualization) of vegetable species and specimens to our knowledge its application to the processing of datasets targeted at lichenometry has never been applied. 

This work presents a set of image acquisition and processing tools and methodologies to solve the problems associated with data acquisition and processing for lichenometry application, namely the individualization and measurement of lichens (percentage of coverage, number of individuals and size of each individual in $mm^2$). 
The proposed solution includes methodologies and tools for image perspective correction and scaling, and computer vision, machine learning algorithms to automate the segmentation, individualization, and measurement of lichens in images. 
The objective is to reduce data collection time and ease field work (e.g., selection of thalli to be measured), to improve accuracy in the individualization and measurement of thalli (analysis and processing of collected data) and to create a methodology that can be replicated in various environments (lichens and rock surfaces) and by any user.

\section{Applied techniques}
This section presents the various techniques applied in the work here described. 

The automatic classification of images requires the definition of training data-sets. 
This task should be done by a operator that can identify and distinguish in a image a lichen from its background. 
Currently this is already done using commonly available image processing tools such as Photoshop, but is time consuming. 
To help the operator perform the trains set creation task the GrabCut algorithm was selected. 
Although still requiring an operator to select areas (lichens or background) on a image, it automatizes the definition of large areas that contain lichens or background. 

Image classification can follow a pixel-based approach, where pixels with similar characteristics belong to the same class, or an object-based approach where groups of pixels need to be aggregated into objects. These objects will be classified and represented as a set of features. To define these objects in the images we use the Simple Linear Iterative Clustering (SLIC) technique.

For the classification of the images in the data-set an automatic mechanism should be implemented. From the evaluation of the various possibilities applicable to the problem the use of SVM and Random forest techniques were selected.

\subsection{Graph Cut}

To help operators to define the pixels that correspond to lichen or background the GrabCut \cite{937505, 10.1145/1015706.1015720} was used. 

Grabcut is a segmentation technique that classifies a pixel as background or foreground by building a graph with a certain energy function and defining a cut \cite{boykov2001experimental} between two special nodes previously labeled as background or foreground. 

To calculate a representation of the two areas, the user defines a set of seed pixels that either are part of the foreground or background.  
These pixels can be marked with a rectangle that completely encloses the background or by a set of line segments that either overlap foreground or background pixels, without crossing boundaries.  
These user defined pixels are used to calculate the foreground and background color distribution models, such as the intensity histograms or \cite{937505} a Gaussian Mixture Model \cite{10.1145/1015706.1015720} ({GMM}). 
With a model that describes the seed pixels it is possible to build a graph that connects other pixels to a source and a sink (two pixels previously marked by the user as foreground or background) where each edge has an energy cost that represents the likelihood of the pixel to be represented by one of the models. 

By applying a cut algorithm, it is possible to determine whether each pixel belongs to the foreground (less cost to the source pixel) or to the background (less cost to the sync). 
A Min-cut/Max-Flow algorithm is used to segment the graph. 
This algorithm determines the minimum cost cut, calculated by the sum of all the weights of the links that are cut (minimization of the energy/cost function), which will separate the Source and Sink nodes. 
Once the Source and Sink nodes are separated, all pixel nodes connected to the Source node become part of the foreground, and the rest become part of the background.
 
After classification the estimates can be further corrected.
The user can draw new background or foreground segments, thus intermediately  correcting the classification on those pixels, and creating a new model, and running the optimization.

\subsection{Simple Linear Iterative Clustering (SLIC)}
 \label{slic-section}

If images contains pixels belonging to the same class but having a high variance in the used spectral bands (red, green and blue, multi or hyper-spectral bands) there are  heterogeneous features with complex patterns, and  an object-based approaches provide better results \cite{aghababaei2021classification,akar2012classification}.
To our understanding applying object-based image classification to lichens, due to the remarkable complexity of the vegetative plant body that comprises the thallus,  is more efficient. 

Adding to the natural color variation between lichens of the same species,  saxicolous lichen (lichens that grow on rocks) thalli, for example, are frequently heterogeneous comprising several distinctive features, such as the areola, the prothallus, and apothecium \cite{doi:10.1177/030913338500900202}.
In these object-based approaches the optimal size of segments depends with image resolution, size of the objects in the image, and the variety of both parameters within an image data-set, bringing additional challenges to object-based image classification \cite{kim2008estimation}. 

To handle the creation of these objects (areas that should be considered lichen or background) we selected the use of Simple Linear Iterative Clustering (SLIC).
{SLIC} \cite{slic_powaaa} creates super-pixels based on k-means clustering. Super-pixels are small regions of pixels in the image that share similar properties (color). Super-pixels simplify images with a large number of pixels making them easier to handle in many domains (computer vision, pattern recognition and machine learning).
This algorithm generates super-pixels by grouping pixels based on their color, similarity and proximity in the image plane.

The {SLIC} function from the python library skimage has the following parameters:

\begin{itemize}
    \item n\_segments: Approximate number of {SLIC} segments created for the input image.
    
    \item compactness : Balances color proximity and space proximity. Higher values give more weight to the proximity to space, making super-pixel shapes more square. This parameter strongly depends on the contrast of the image and the shapes of objects in the image.
    
    \item sigma: Width of the Gaussian smoothing kernel for the preprocessing of each dimension of the image. The same sigma is applied to each dimension in the case of a scalar value. Zero means no smoothing.
    
\end{itemize}

It was decided to divide the images into regions given by {SLIC} to train the classifiers. Preliminary tests carried out in the course of this work indicate that this approach provides good delimitation between lichens and rock, i.e. the boundary regions of the {SLIC} segments largely coincide with the boundaries between lichens and rock.
This drastically reduces the number of features (number of features becomes equal to the number of {SLIC} segments instead of the total number of pixels).

\subsection{SVM and Random Forests}

In machine learning, support-vector machines ({SVM}s) are supervised learning models with associated learning algorithms that analyze data for classification and regression analysis \cite{hastie_09_elements-of.statistical-learning}.
Given a training data set consisting of already classified images, a {SVM} training algorithm builds a model that assigns new data to one class or another, making it a non-probabilistic binary linear classifier. The {SVM} maps training data to points in space and seeks to maximize the width of the interval between the points of the two classes (in the case of binary classification there are only two classes). New data is then mapped into that same space and predicted to belong to one of the two classes based on which side of the boundary they lie on. In addition to performing linear classification, {SVM} can efficiently perform non-linear classification using what is called the kernel trick by implicitly mapping its inputs into high-dimensional feature spaces.

Formally, a {SVM} constructs a hyperplane or set of hyperplanes in a high-dimensional space, which can be used for classification, regression, or other tasks such as outlier detection \cite{scikit-learn}. Intuitively, in the case of classification, good separation is achieved by the hyperplane that has the greatest distance to the nearest training data point of any class, since in general the larger the margin, the smaller the generalization error of the classifier \cite{hastie_09_elements-of.statistical-learning}.

While the original problem can be defined in a dimensionally finite space, it is often the case that the sets to be discriminated are not linearly separable in that space. For this reason, it has been proposed that the original space be mapped to a much higher space, presumably facilitating separation in that space \cite{Boser92atraining}. To keep the computational load reasonable, the mappings used by the {SVM} schemes are designed to ensure that the scalar products of the input data vector pairs can be computed easily in terms of the variables in the original space by defining them in terms of a kernel function $\displaystyle{k(x,y)}$ selected to fit the \cite{press2007numerical} problem.

Random Forests is a particular type of learning algorithm based on decision trees.
In decision tree learning, decision trees are used as predictive models to go from observations about an item (represented in the branches) to conclusions about the target value of the item (represented in the leaves).

Tree models where the target variable can take a discrete set of values, are called classification trees; in these tree structures, the leaves represent the class labels and the branches represent conjunctions of features that lead to those class labels.

Some techniques, often called ensemble methods, build more than one decision tree. 
Bootstrap aggregated (bagged) decision trees build multiple decision trees by repeatedly resampling training data with replacement, and voting the trees for a consensus prediction \cite{Breiman1996}. 
A random forest classifier is a specific type of bootstrap aggregating. 
Random forests correspond to an ensemble learning method for classification, regression and other tasks, which works by building a multiplicity of decision trees in the training period. 
For classification tasks, the result of random forests is the class selected by most trees \cite{hastie_09_elements-of.statistical-learning}.

\section{Automating image processing}
\subsection{Description of the system from the user's point of view}

The objectives of this project are related to the development of a tool for automatic measurement of lichen dimensions in order to contribute to better data acquisition for dating with lichenometry in geological and/or archaeological studies.

\begin{figure}[htb]
\centering
\includegraphics[width=0.2\textwidth]{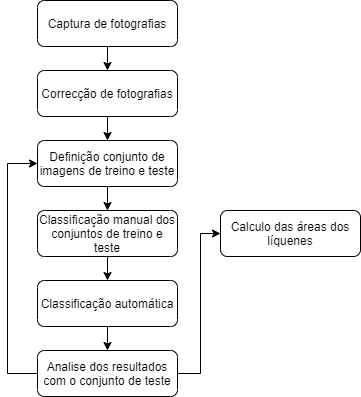}
\caption{User flow.}
\label{fig:fluxo user}
\end{figure}

The user flow (\Cref{fig:fluxo user}) consists of the following steps:

\begin{itemize}
    \item Photo Capture - The tool will be applied to previously sampled data sets or to new data sets acquired during the course of this project. It is up to the user to choose which data set to analyze using the program.
    \item Photo correction - Given the chosen data set, each photo in the data set is corrected for perspective errors.
    \item Definition of the set of training and test images - Definition of two subsets (training and test) of images belonging to the chosen data set. The training images serve to train the classifier to identify the regions of interest (lichens) and the test images serve to evaluate and benchmark the performance of the system.
    \item Manual classification of the training and test sets - The regions of interest (lichens) of the training and test images are explicitly identified by manually classifying these images with the help of a background extraction tool. 
    This results in binary images with the lichen class regions in white and the rest in black. 
    These images together, with the corresponding training and test images, allow the program to learn.
    \item Automatic classification - The program will automatically classify the regions of the data set images that are lichens.
    \item Analyzing results with the test set - The system returns the segmented images and the performance of the measurements made on the test set of images. Depending on the performance it may be necessary to re-run the program with more training images.
    \item Calculation of lichen areas - For each chosen segmentation the area of each individual lichen is measured.
\end{itemize}

The next sections describe each element present in \Cref{fig:fluxo user} in detail.

\subsection{Images capture and correction} \label{3.2 miras}

One of the problems associated with photograph sampling, particularly relevant when the goal is to extract spatial information (such as the area occupied by a lichen thallus), is image deformation. This occurs either due to the oblique orientation of the photograph relative to the surface, or due to the deformation of the lens.

Thus, in the acquisition of new images in the field, we propose the use of 4 blue targets arranged over the vertices of a square/rectangular region that includes the area of interest to be photographed. This will allow corrections and transformations to be applied in order to compensate for perspective errors and assigns a scale.

The size (width and height) of the rectangular region enclosed by the targets must be recorded during sampling and later entered into the program so that image deformation can be removed.

In a first iteration of the target system, the detection of the Leica® topographic targets was made using {SIFT} \cite{790410} descriptors and using  {RANSAC}.
The  {SIFT} generates descriptors that represent notable points in an image allowing matching the {SIFT} descriptors of the search image (image with only the topographic target to detect) to the {SIFT} descriptors of the distorted image (the matching will be done with one of the 4 targets in the image). 
This correspondence is done with {RANSAC}, knowing that a correspondence can only exist between points of the same plane (homography). 
Since all 4 targets were identical, whenever a correspondence was made between the search image and a target, the program had to cover that specific target from the image. 
Otherwise the correspondence would always be made between the same descriptors {SIFT}, belonging to the same target.

While this method produces some promising results in controlled experiments, it is not robust enough to perform target detection in images with feature-rich rocky backgrounds. More importantly, when the perspective deformation of the image is more intense this system can no longer detect all 4 targets and starts matching wrong points. 

Robustness was improved using a simpler solution through color segmentation, using 4 circular targets with the same blue color.
This color was chosen in order to provide a good contrast and because it is uncommon in both rock formations and lichens.

In order to determine the parameters necessary for color segmentation of the targets, the images photographed with the target system were first converted from {RGB} to {HSV}. This makes it easier to define the upper and lower limits of the color channels in order to segment the targets. The experimentally defined H, S and V ranges were: H[95-105], S[85-255] and V[170-245].

The system works by detecting the centers of mass of the segmented regions, corresponding to the targets. 
With the coordinates of these 4 points and knowing the real geometry of the targets (rectangle with 27.2cm length and 18.5cm height, for example) a geometric transformation matrix is calculated. 
With this matrix, the perspective of the image is transformed in order to eliminate errors from the original perspective and facilitate measurements.

\begin{figure}[htb]
\centering
    \subfigure[]{\label{fig:mira_A_1}\includegraphics[width=0.24\columnwidth]{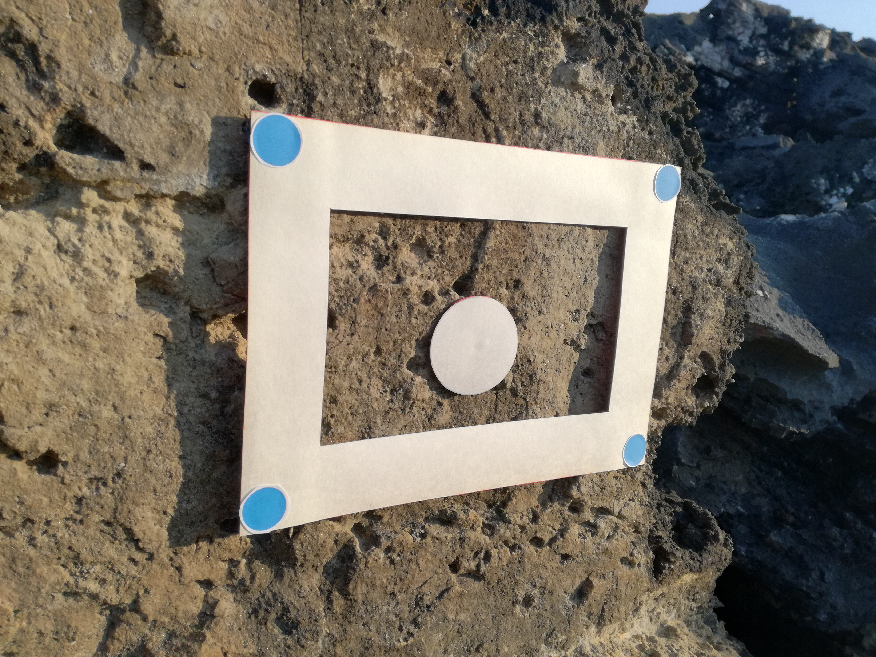}}
    \subfigure[]{\label{fig:mira_B_1}\includegraphics[width=0.24\columnwidth]{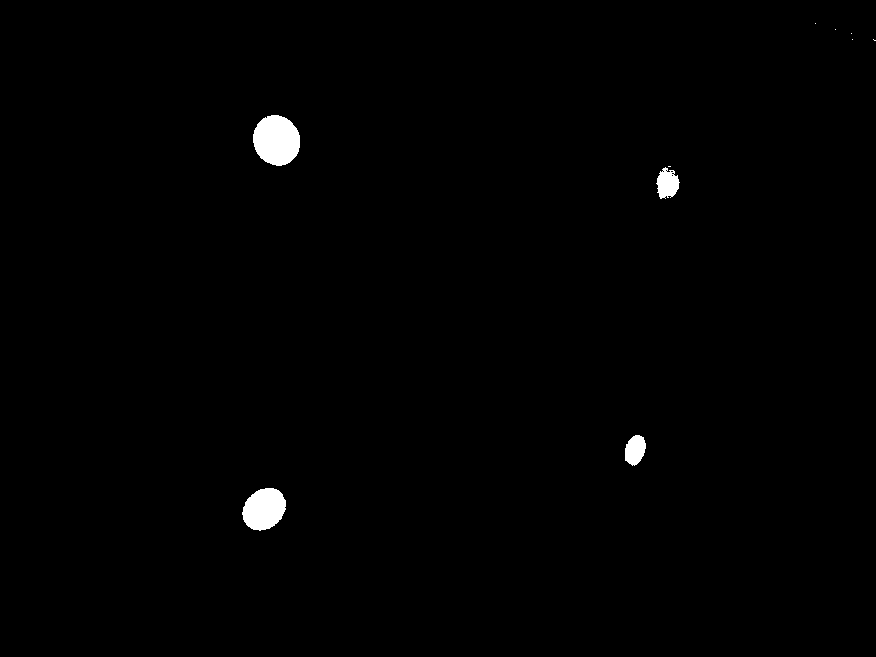}}
    \subfigure[]{\label{fig:mira_C_1}\includegraphics[width=0.24\columnwidth]{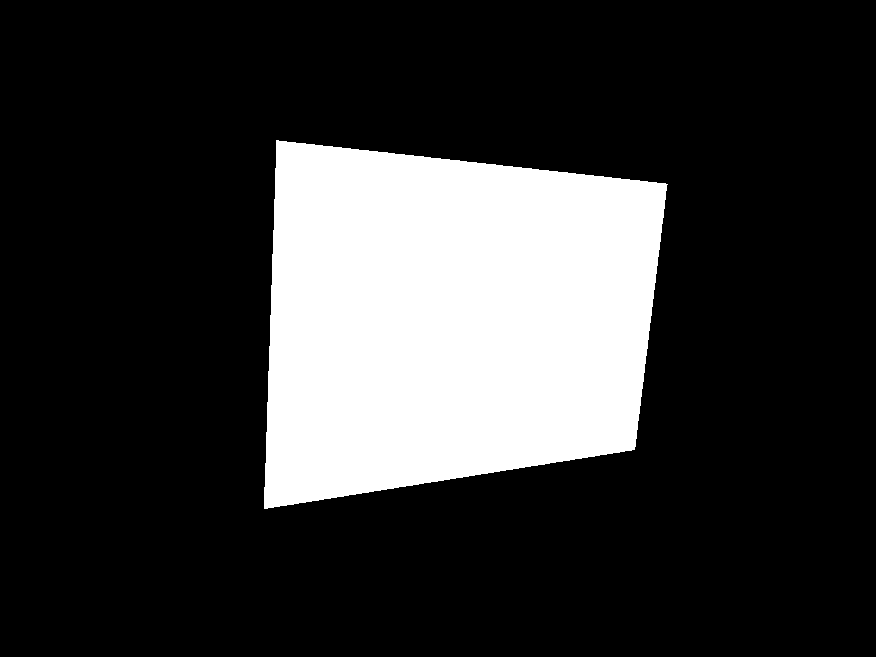}}
    \subfigure[]{\label{fig:mira_D_1}\includegraphics[width=0.24\columnwidth]{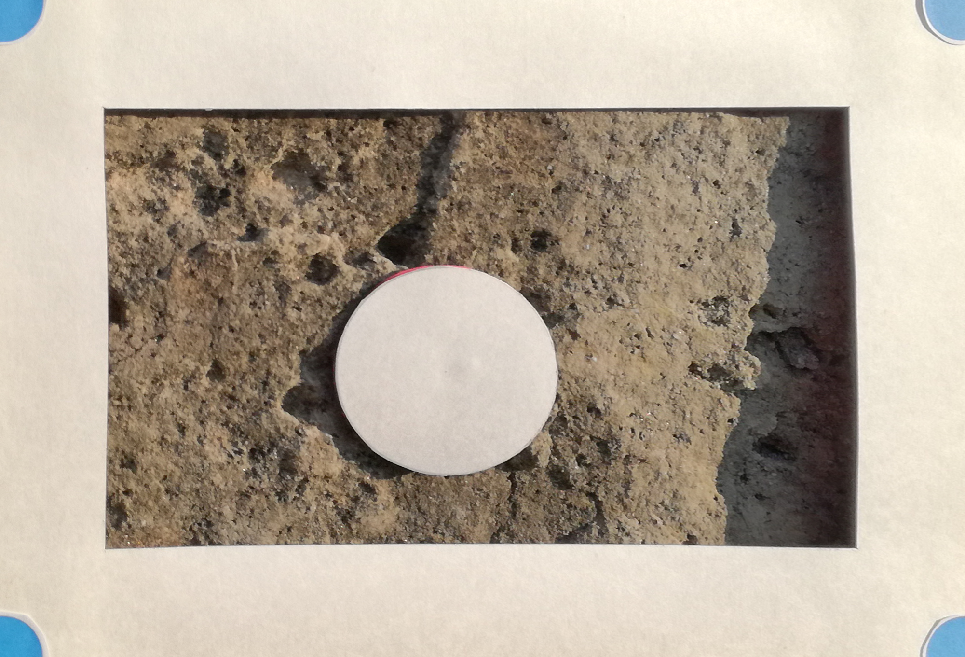}}
    \caption{Image correction example: (a) original photograph, (b) target detection, (c) interest area crop, (d) final corrected image.}
    \label{fig:miras}
\end{figure}

\Cref{fig:miras} shows that the deformation of the circular target in the center of the target region is corrected.

\subsection{Manual classification process} \label{manual-class}

The manual classification generates binary outputs corresponding to the training and test images, in which the regions belonging to the lichen class are identified in white and the others in black, allowing the user to 'teach' the program which features of interest in the images. These binary images, together with the corresponding training and test images, serve as a reference for the program to learn.

The manual image classification process is done using the GrabCut algorithm.
From the user's point of view, the GrabCut algorithm works by accepting an input image where: 
\begin{itemize}
\item the user identifies several areas corresponding to the lichens,
\item the user identifies several areas corresponding to the background,
\item the system updates the binary result and,
\item if the resulting classification contains obvious errors, the user can identify additional lichens and background.
\end{itemize}

Sometimes the segmentation performed by the algorithm based on the user's delimitation is far from ideal. In such cases, fine touches need to be made by selecting defective results and marking them properly.

\begin{figure}[htb]
\centering
    \subfigure[]{\label{fig:graph_A}\includegraphics[width=0.32\columnwidth]{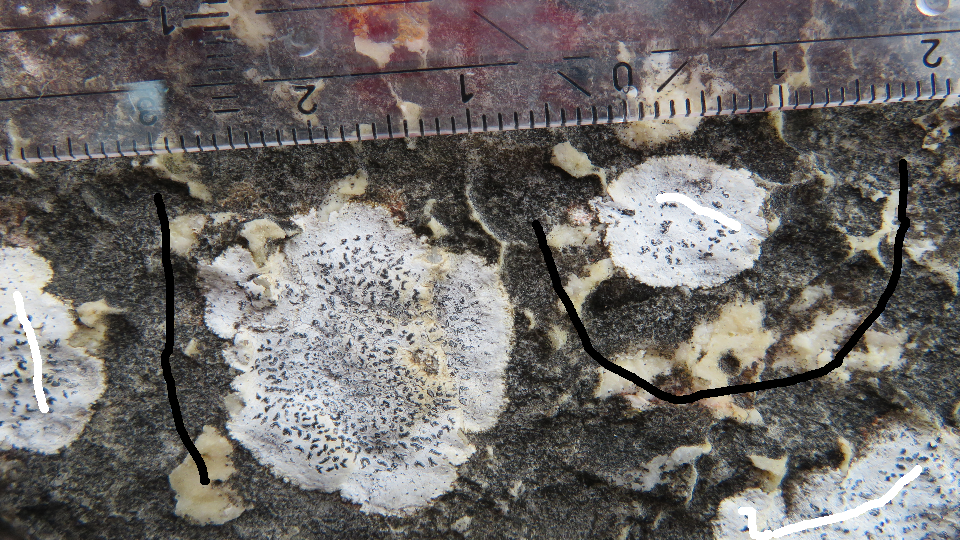}}  
    \subfigure[]{\label{fig:graph_B}\includegraphics[width=0.32\columnwidth]{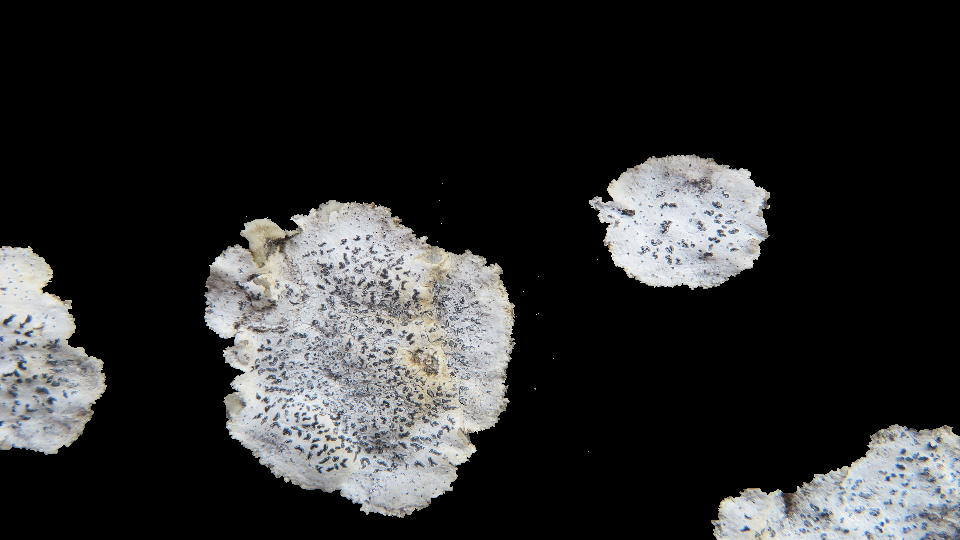}}   
    \caption{Example of foreground (white strokes) and background (black strokes) selection and final segmentation using GrabCut.}
    \label{fig:grabcut}
\end{figure}

As can be seen in the \Cref{fig:grabcut}, after some final retouching, identified by the white (denoting foreground) and black (denoting background) strokes, a good segmentation result is obtained which, in this case, separates lichen from rock.

\subsection{Automatic classification}

In this section all the steps and operation of the components that perform the automatic classification are explained.

\begin{figure}[htb]
\centering
\includegraphics[width=\columnwidth]{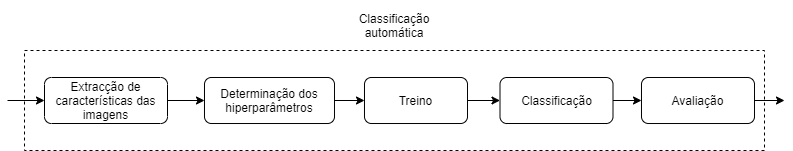}
\caption{Flow of the automatic classification.}
\label{fig:fluxo prog}
\end{figure}

The automatic lichen classification program receives as input data the directory where the images are located (data set). 
The user has to define the number of pictures that will be used for training and also for testing (randomly chosen). 
These pictures will be manually classified as described in \Cref{manual-class}. These data sets allow the classifiers to be trained with properly classified data and provide a benchmark for evaluating the program's automatic classification performance. The flow of the program is illustrated in \Cref{fig:fluxo prog}.

The automatic classification component receives the training and test images as well as the corresponding binary images manually classified with GrabCut. The output returns the segmentations of the remaining images in the data set (images that do not belong to the training and test sets).

The program uses as features the relative frequency histograms corresponding to the {SLIC} segments of the images. 
Alternatively, each pixel could be used as a feature for training and classification. 
The use of the {SLIC} segments allows speeding the execution of the training and classification since the number of training features is reduced (the number of pixels is much larger than the number of {SLIC} segments). 
It also allows preserving some local information of the image regions (fundamental for segmentation), since a pixel itself has no information about the surrounding region (unlike {SLIC} segments).

For each {SLIC} segment, the program will define a relative frequency histogram, representative of the pixels of that segment (in percentage of occurrence of each pixel of each color). Each relative frequency histogram serves as features to train the classifiers.

For each image, the program generates different sets of {SLIC} segments, each created with a different set of parameters. The range of {SLIC} parameters tested is: n\_segments = [2000, 1000, 500], compactness = [20, 10], sigma = [3, 1], threshold = [0.5].

\begin{figure}[htb]
\centering
    \subfigure[]{\label{fig:SHOW_A}\includegraphics[width=0.32\columnwidth]{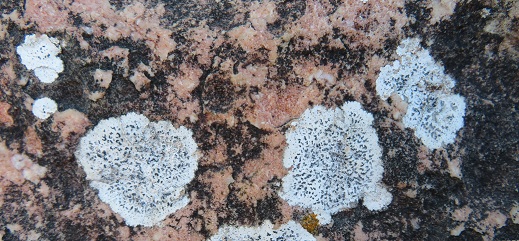}}
    \subfigure[]{\label{fig:SHOW_B}\includegraphics[width=0.32\columnwidth]{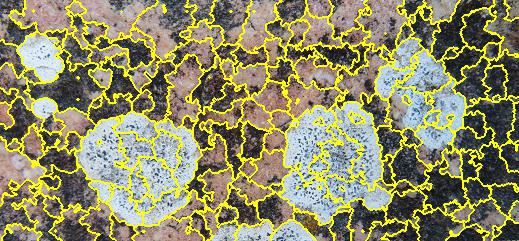}}
    \subfigure[]{\label{fig:SHOW_C}\includegraphics[width=0.32\columnwidth]{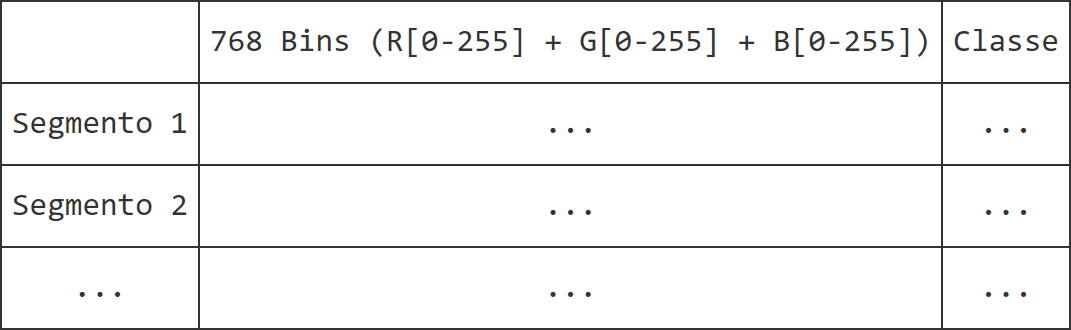}}
    \caption{Extraction of features from images.}
    \label{fig:show}
\end{figure}

Thus, the training and test sets are converted into 12 different sets of training and test data corresponding to feature extraction with the 12 possible combinations of {SLIC} parameters.
Each data set (training, test or singular image to segment), is therefore represented by a table where the rows correspond to the {SLIC} segments of all the images in the set and the columns correspond to the pixel color frequency histograms of each {SLIC} (row). 
In the case of the training and test sets, there is also a last column representing the class of each segment  (lichen or background), product of the manual classification. 
\Cref{fig:show} shows the feature extraction for an image.

Both the {SVC} classifier and the RandomForestClassifier have various combinations of parameters that can affect the classification performance:
\begin{itemize}
    \item {SVC} - 'C': [1,10,100], 'kernel': ['rbf', 'linear', 'poly'], 'degree': [2,3,4,5], 'gamma': ['scale', 'auto'], 'max\_iter': [500, 1000]
    \item RandomForestClassifier - 'n\_estimators': [150, 100, 50], 'criterion': ['gini', 'entropy']
\end{itemize}

In order to find the best combination of parameters, it is necessary to make an initial evaluation usually called hyperparameter estimation.

For this, cross-validation using 5 folds was applied to the 12 sets derived from the {SLIC} parameters.

Training is done with the multiple combinations of hyperparameters of the classifiers and for each combination {SLIC} their performance is obtained.

The combination of parameters (for {SVC} and RandomForestClassifier) with the best performance is used in training and further classifications.

After determining the hyperparameters to be used by each of the classifiers, 24 classifiers are trained using the training set. These 24 versions correspond to instantiations of the {SVC} and RandomForestClassifier classifiers with the previously defined hyperparameters combined with 12 configuration alternatives of {SLIC}.

To evaluate the performance, the program segments the 12 test sets with the two classifiers thus producing 24 segmentations for each test set image.

The metric chosen to perform this evaluation was the  Matthews correlation coefficient\cite{MATTHEWS1975442} (MCC). Each model segments the images in its test set and assigns to each a {MCC} value using manual classification as a reference.
The performance of each classifier/parameter is obtained by averaging the {MCC} for the test set.
This performance indicator for each of the 24 combinations of classifiers and parameters allows the to decide on adding more images to the training set and repeat the training and classification process;
or choose one of the combinations and use the corresponding classification results.

Once trained and evaluated, it is possible to choose the classifier that can produce the best results based on the {MCC} calculated earlier. This classifier processes the remaining images in the data set and produces the corresponding classifications.

\subsection{Calculation of lichen areas}

The results of the classifications  identify the areas occupied by lichens. However, it is essential that other data can be  extracted, such as the  number and area of lichens thalli.

These calculations are performed by a program that receives the binary images from the segmentations (black and white images) and, for each lichen, assigns an index, returns the area, filled\_area (number of pixels in the region with all holes filled) perimeter and centroid coordinates.

The program uses the \textit{regionprops} function from the \textit{measure} library of \textit{skimage} to perform the analysis of the image regions.

If the user assigns a scale to the image, it is possible to convert the area of each lichen from pixels to mm².

The scale can be taken automatically from the image if it was captured with the developed target system, or manually if the image includes another type of measuring instrument (ruler).

\section{Evaluation}

\subsection{Description of data sets and test environment} \label{sec-4.1}

All code was developed and tested on a machine with the following specifications: Windows 10, 64-bit, 8 GB of RAM, Intel(R) i7-2630QM CPU @ 2.00GHz 4 Cores CPU, Python 3.7.9, Opencv 4.5.1, Pandas 1.2.1, Scikit-learn 0.24.1, Scikit-image 0.17.2, Anaconda 1.9.12, Spyder 5.0.0.

Due to high execution times some tests (automatic classification and feature tests) were done on a different computer: Ubuntu 18.04.4, 15GB of RAM, Intel(R) Xeon(R) E3-1230 v5 @ 3.40GHz 8 Cores CPU.

The data sets present in \Cref{tab:datasets-table} were used to evaluate the automatic classification program.

\begin{table}[h]
\centering
\scriptsize
\begin{tabular}{llcc}
\hline
\textbf{Name} & \textbf{Local} & \textbf{Nº imagens} & \textbf{Resolution} \\ \hline
Terraço & Antartida & 63 &  3888 x 5184 \\ \hline
Nazaré 1 & Nazaré & 27 & 1944 x 2592 \\ \hline
Nazaré 2 & Nazaré & 40 & 3456 x 4608 \\ \hline
Nazaré 3 & Nazaré & 52 & 3456 x 5184 \\ \hline
Muro Castelejo 2 & Fundão & 38 & 3456 x 5184 \\ \hline
Muro Escola Castelejo & Fundão & 17 & 3456 x 5184 \\ \hline
Cascais & Cabo Raso & 63 & 2000 x 3008\\ \hline
\end{tabular}
\caption{Data sets used.}
\label{tab:datasets-table}
\end{table}

\subsection{Images capture and correction}
In order to evaluate the targets system, experiments were performed with 18 photographs taken with the special mark. The experiments were performed at the same place where the Cascais data set was acquired (\Cref{tab:datasets-table}). The procedure is similar to what would be done to capture lichens, but in this case a special mark is used. Each experiment consists of: 
i) placing the special mark on a rocky surface, 
ii) placing the targets on the same surface so that they surround the mark, 
iii) taking a picture of the targets  and mark,
iv) correction of the image, 
v) measuring the major and minor axes of the mark,  and 
vi) comparing the results to the reference values.

Both the targets and the mark have well-defined dimensions. 
The targets correspond to 4 blue circles glued onto a rectangular base, the centers of these targets form a rectangle  272mm long and 185mm wide. The special mark is a white circle with a 60mm diameter. The mark and targets are fixed to the rock surfaces with adhesive paste.

\begin{figure}[htb]
\centering
    \subfigure[]{\label{fig:eli_A}\includegraphics[width=0.32\columnwidth]{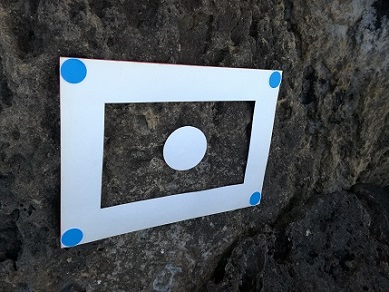}}
    \subfigure[]{\label{fig:eli_B}\includegraphics[width=0.32\columnwidth]{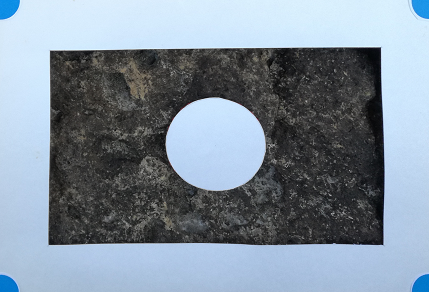}}
    \subfigure[]{\label{fig:eli_C}\includegraphics[width=0.32\columnwidth]{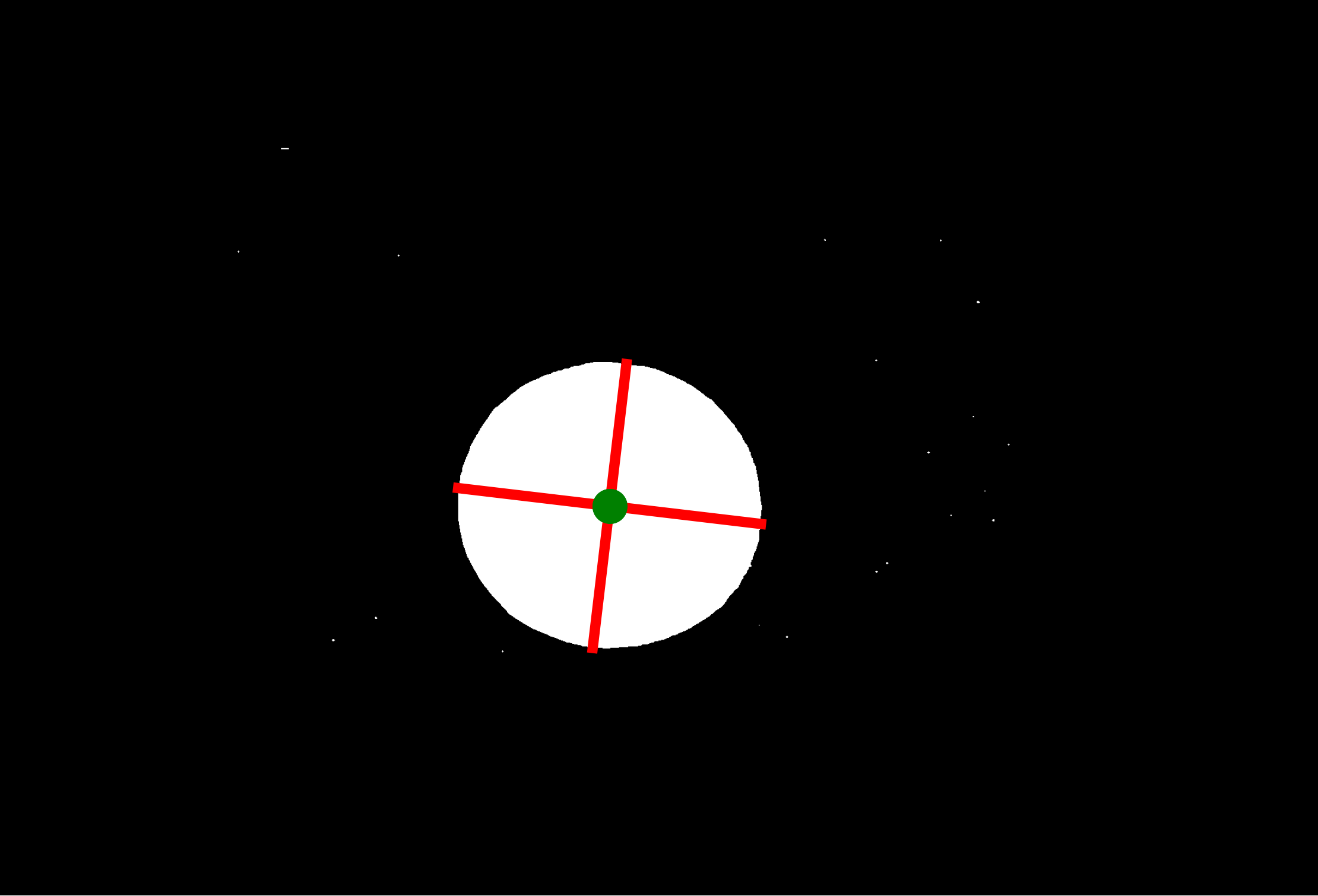}}
    \caption{Example of measuring the major/minor axes of the mark.}
    \label{fig:eli_tests}
\end{figure}

The measurements of the minor and major axes of the mark in the corrected  image (\Cref{fig:eli_B}) were performed automatically using the major\_axis\_length and minor\_axis\_length parameters of the \textit{regionprops} function belonging to the python skimage library (\Cref{fig:eli_C}). 

In one experiment it was not possible to detect the 4 targets, due to the presence of sky in the image, with a similar hue to the targets.

The average error obtained was 7.26\%, and was from the fact that the target and the mark are not exactly in the same plane, since rocks surfaces ares irregular. 
As well as possible measurement errors of the mark radius and distances between the targets.



\subsection{Manual classification}
The use of GrabCut allows the user to reduce the time spent on image classification. This new tool is used for the creation of the training/test image sets, but can also be used alone, replacing the processing that uses other commercial software.

The evaluation of GrabCut's performance was performed by comparing ArcGIS™ and Photoshop® classifications as a reference, focusing on both time and classification. 
To perform this evaluation, a researcher was asked to classify 18 photographs: 9 were classified with Photoshop® and GrabCut and another 9 were classified with ArcGIS™ and GrabCut. 

\begin{figure}[htb]
    \subfigure[ArcGIS™]{\includegraphics[width=0.5\columnwidth]{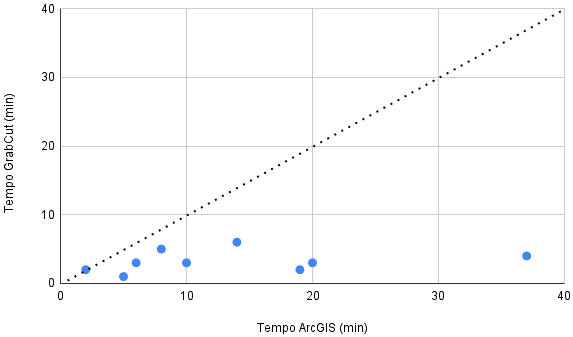}}
    \subfigure[Photoshop®]{\includegraphics[width=0.5\columnwidth]{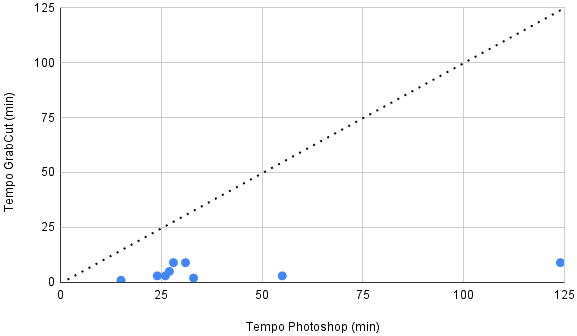}}
    \caption{Time comparison between GrabCut and manual methods.}
    \label{fig:GrabCut_eval}
\end{figure}

It can be seen (\Cref{fig:GrabCut_eval}) that GrabCut is always faster than the methods using Photoshop® and ArcGIS™. 
Furthermore, in a series of experiments, a 90\% gain was obtained. In other words, in these cases the user only needs to spend 10\% of the time that would have been spent using commercial software.

The associated errors obtained for all experiments were analyzed by comparing the GrabCut results against the results obtained with Photoshop® and ArcGIS™, chosen as references. 
This comparison was made in terms of {MCC}.

\begin{figure}[htb]
\centering
\includegraphics[width=0.5\columnwidth]{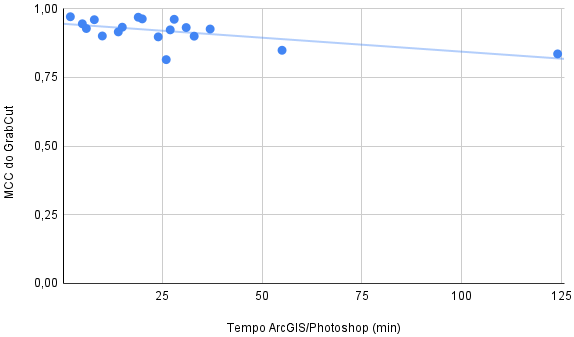}
\caption{Evolution of GrabCut error with manual classification time.}
\label{fig:MCC_tempo}
\end{figure}

A mean {MCC} of 0.9174 was determined (y-axis of \Cref{fig:MCC_tempo}). 
In order to analyze the {MCC} results in more detail, the relationship between {MCC} and the ArcGIS™/Photoshop® manual classification time is shown in \Cref{fig:MCC_tempo}.
It can be seen that there is a trend, and that the more complex/difficult is to segment an image, the greater the error of the classification performed with GrabCut.
So, even in complex images, for which manual classification took approximately 2 hours, the GrabCut classification was under 10 minutes and the MCC was still above 0.75.

\subsection{Automatic classification}




\subsubsection{Feature creation}

Since {SLIC} does clustering of areas in the images, it introduces errors, as the boundaries of the regions given by {SLIC} may not completely coincide with the lichen borders.
Meaning that a given {SLIC} segment may contain both background and lichen pixels.
Due to this possibility, an evaluation was be performed to quantify the error associated with the {SLIC}.

This evaluation was based on  the same 18 images from the previous section and the corresponding (manually performed) binary classifications.
However, they were downscaled to different resolutions (40\%, 35\%, 30\%, 25\%, 20\%, 15\% and 10\%) from the original images in order to speed up the evaluation process and to study the temporal scalability of these features.

The {SLIC} parameters tested were the same as those used in the automatic classification: n\_segments = [2000, 1000, 500], compactness = [20, 10], sigma = [3, 1], threshold = [0.5].

All 18 images were tested for the 12 parameter combinations {SLIC}, i.e. 216 tests and these tests were repeated for the same images but at different resolutions.

In each test, the SLIC clustering was applied to produce a set of segments. Each segment was assigned one class with help of the corresponding classification done with Photoshop®/ArcGIS™.

It is important to remember that the threshold parameter of the {SLIC} enters in the class assignment to each segment. 
For example, for a value of 0.5 (50\%), when a given segment contains more than half of lichen pixels (you can tell by comparing with the classification done manually), then the lichen class is assigned to that segment, otherwise the background class is assigned.

Finally, the differences were compared and the times were measured. Feature generation times are the sum of the time to create the {SLIC} segments, the time to convert each {SLIC} segment into the corresponding relative frequency histogram (according to the colors of the pixels in that region) and, in the case of training data, the time to assign the class to each segment.

The average values of precision and of the {MCC} attained are relatively constant for the different image resolution, and the precision always in the order of 98\% and {MCC} of 0.87.
In cases where the size of the segments (given by the parameter n\_segments) is larger than the lichen regions, the program may not be able to convert the lichen regions in the images to their features, originating  feature segments assignment only to the background class. 
The precision value in these cases may be high (corresponding to the percentage of the background present in the image), however the {MCC} metric detects these cases resulting in {MCC}=0.

The experiments also show that the times depend exponentially with image resolution since, the higher the resolution of the images (more pixels to be processed), the longer the time it takes to calculate the histograms. 
And the higher the number of {SLIC} segments is, the longer it takes to calculate the histograms (each {SLIC} segment is converted into a corresponding histogram). It should be noted that the calculation of the histograms is the most time consuming process.

However, the most important thing to consider in these results is the error of the conversion to {SLIC} segments. 
We want to minimize this error since  these features based on {SLIC} segments are used to train the classifiers. 
This error is due to the fact that the borders of the {SLIC} regions do not always coincide with the  lichen borders.
In the case of the tests performed on the set of images reduced to 30\%, it was found that, in general, the error is low (high {MCC} values) and that this particular distribution presents a mean of 0.8679 and a standard deviation of 0.1327.

\subsubsection{Learning curves and system scalability}

In order to measure the learning capacity and scalability of the algorithms, several training cycles were performed for each of the 7 sets of images, each with an incremental number of training images. 
At the end of each training cycle, the images from the test set are segmented and compared to the reference set (manual classifications with GrabCut) to verify the cycle quality of the segmentations (with the metric {MCC}). Scalability was studied by analyzing the execution time of each cycle.

As described in the previous sections, in each training cycle, the two classifiers (RandomForestsClassifier and  {SVC} were trained for each combination of the 12 {SLIC} parameters, resulting  in 24 training cycles. 
Due to the characteristics of this evaluation, which implies the execution of several training cycles and that in each one the number of training images is increased (which further increases execution times) it was decided to exclude cross-validation. This was necessary to keep execution times at reasonable values.

The default parameters  used for RandomForestsClassifier were: 'n\_estimators': [100], 'criterion': ['gini'],and for the  {SVC} were: 'C': [1], 'kernel': ['rbf'], 'gamma': ['scale'], 'max\_iter': [-1]

\begin{figure}[htb]
\centering
    \subfigure[]{\label{fig:1}\includegraphics[width=0.24\columnwidth]{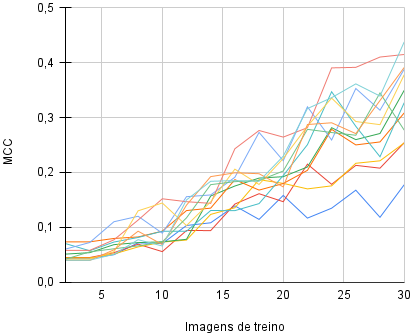}}
    \subfigure[]{\label{fig:2}\includegraphics[width=0.24\columnwidth]{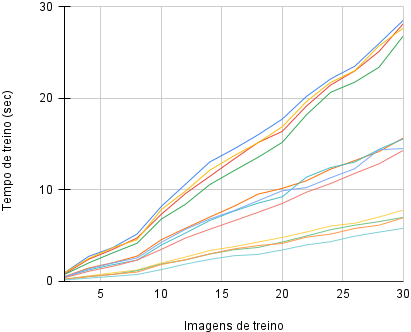}}
    \subfigure[]{\label{fig:3}\includegraphics[width=0.24\columnwidth]{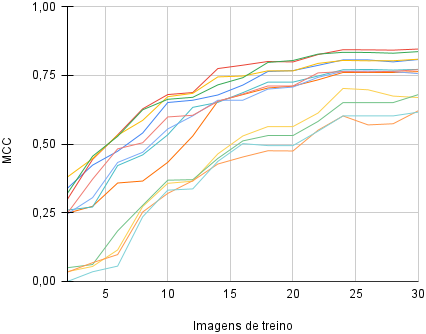}}
    \subfigure[]{\label{fig:4}\includegraphics[width=0.24\columnwidth]{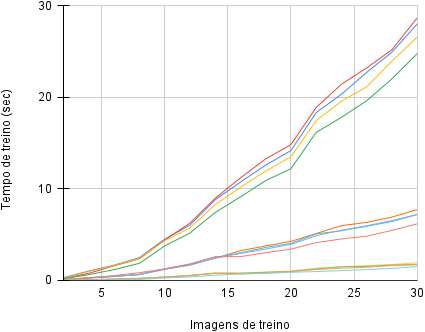}}
    \subfigure[]{\label{fig:5}\includegraphics[width=0.24\columnwidth]{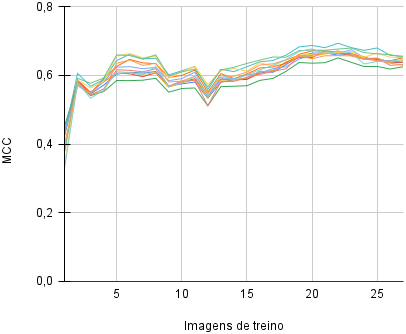}}
    \subfigure[]{\label{fig:6}\includegraphics[width=0.24\columnwidth]{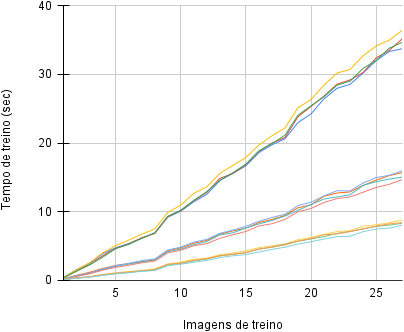}}
    \subfigure[]{\label{fig:7}\includegraphics[width=0.24\columnwidth]{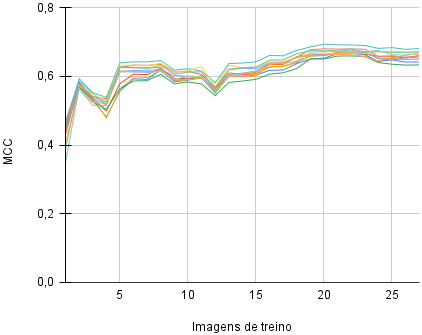}}
    \subfigure[]{\label{fig:8}\includegraphics[width=0.24\columnwidth]{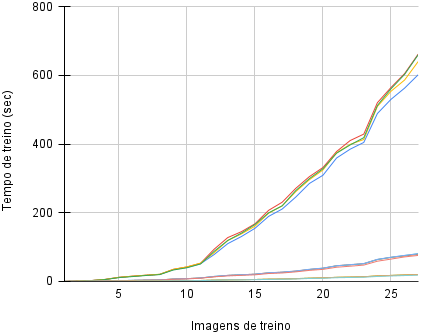}}
    \caption{Learning and scalability curves for Antarctica (a, b, c, d) and Cascais (e, f, g, h) datasets.}
    \label{fig:learn_curves}
\end{figure}

\Cref{fig:learn_curves} illustrates the results obtained for the Antarctica and Cascais image sets. Each graph holds information for the 12 possible combinations of  {SLIC} parameters for an individual classifier. 

\Cref{fig:learn_curves} shows that the learning curves tend to increase with the increase in the number of images used in each training cycle.
meaning, the more images are used in training, the better the quality of the classifications.

Comparing the various learning curves shows that, depending on the set of images, the {SLIC} parameters may or may not impact the quality of the segmentations. 
This is due to the intrinsic variance in each set of images. 
It is also important to note that the classifiers  may have differences in the quality of the segmentations , depending on the image set.. 
In general, {SVC} seems to show better results than the RandomForestsClassifier.

Similarly, the training time for both classifiers increases with the number of training images, i.e., the more images used, the longer it takes to perform the training.

Training times show similarities in the learning curves sharing the same parameter {SLIC} n\_segments. This means that the training time is strongly linked to this parameter and that the more segments {SLIC} were generated, the greater the number of features in the training data, which forces the classifiers to process more information, increasing the training time and worsening the scalability of the algorithms.

It can also be seen that the training times of RandomForestsClassifier classifiers evolve linearly with time unlike {SVC} that evolve exponentially.
 We can also see that, for the Cascais image set, the {SVC} classifier has longer training times than the RandomForestsClassifier. Therefore, the RandomForestsClassifier presents better scalability than the {SVC}.

\subsubsection{Segmentation analysis}

In this section we analyze the  segmentations to illustrate some problems and limitations of the developed program , as well as the potential related to this type of segmentation.

Some segmentations were chosen for analysis  to illustrate the characteristics previously enunciated. 
The goal was to evaluate the quality of the produced segmentations, specifically with regard to true/false positive/negative.

\begin{figure}[htb]
\centering
    \subfigure[]{\label{fig:1a}\includegraphics[width=0.1\textwidth]{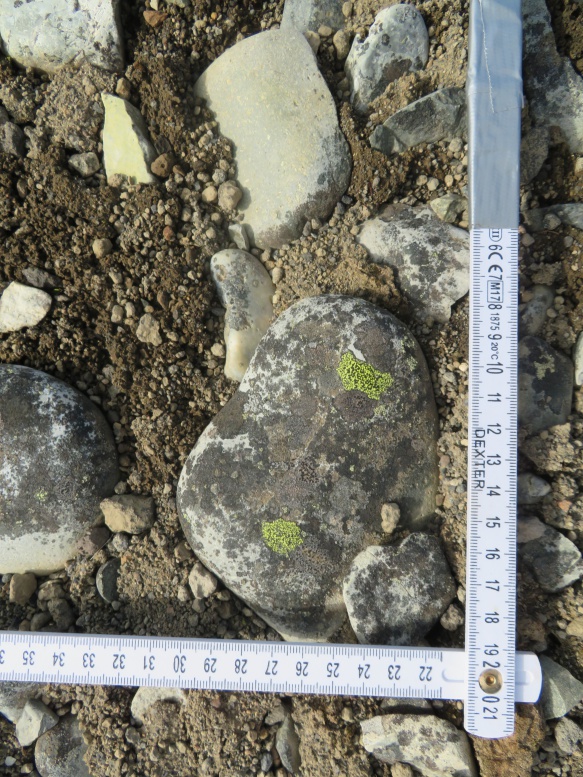}}
    \subfigure[]{\label{fig:2a}\includegraphics[width=0.1\textwidth]{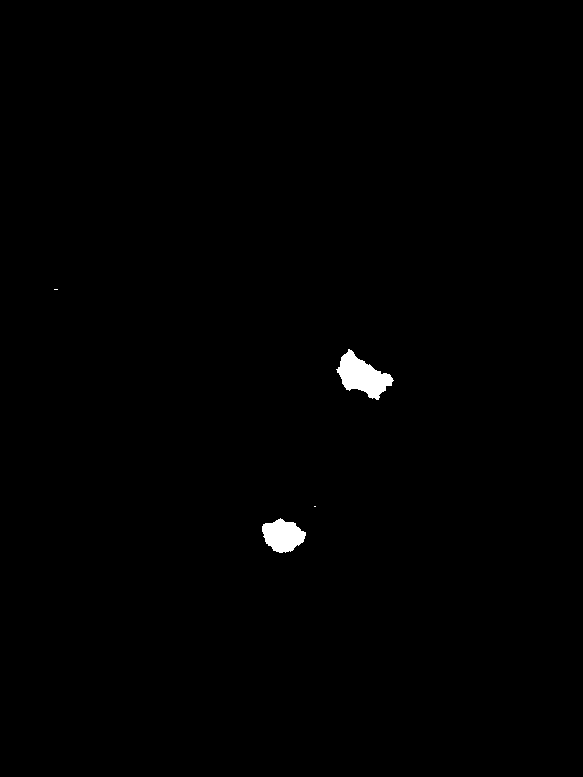}}
    \subfigure[]{\label{fig:3a}\includegraphics[width=0.1\textwidth]{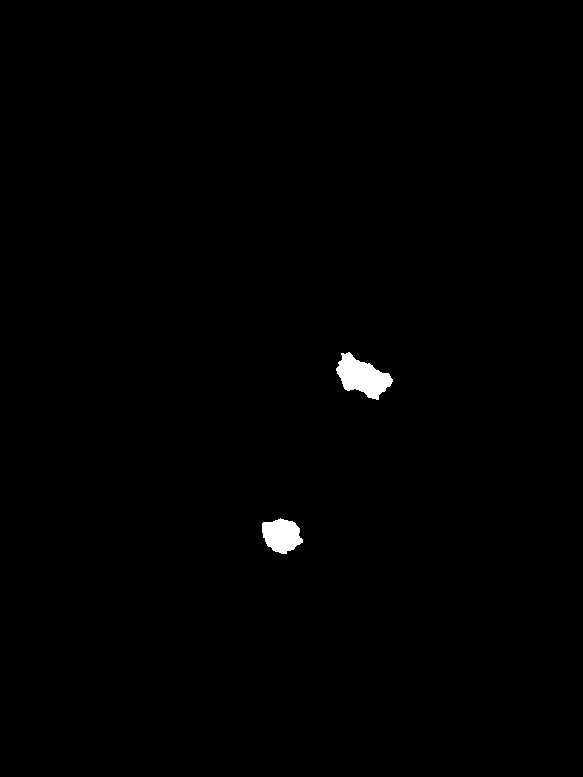}}
    \break
    \subfigure[]{\label{fig:4a}\includegraphics[width=0.1\textwidth]{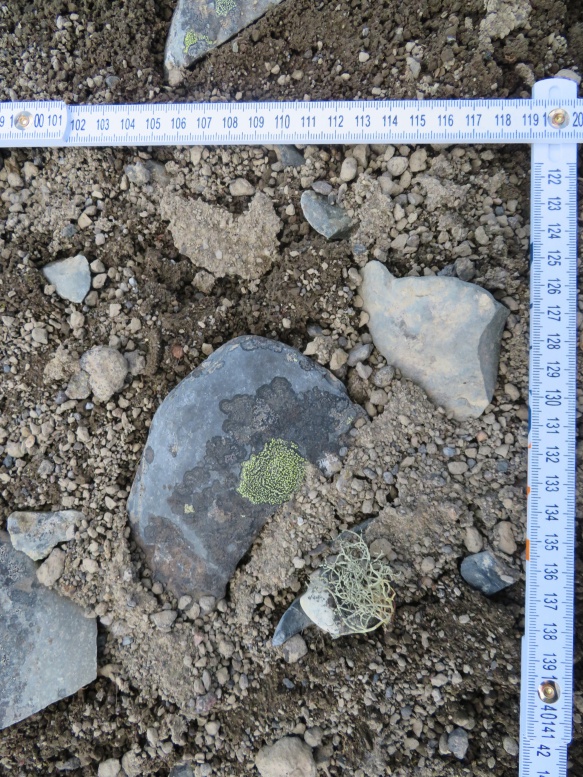}}
    \subfigure[]{\label{fig:5a}\includegraphics[width=0.1\textwidth]{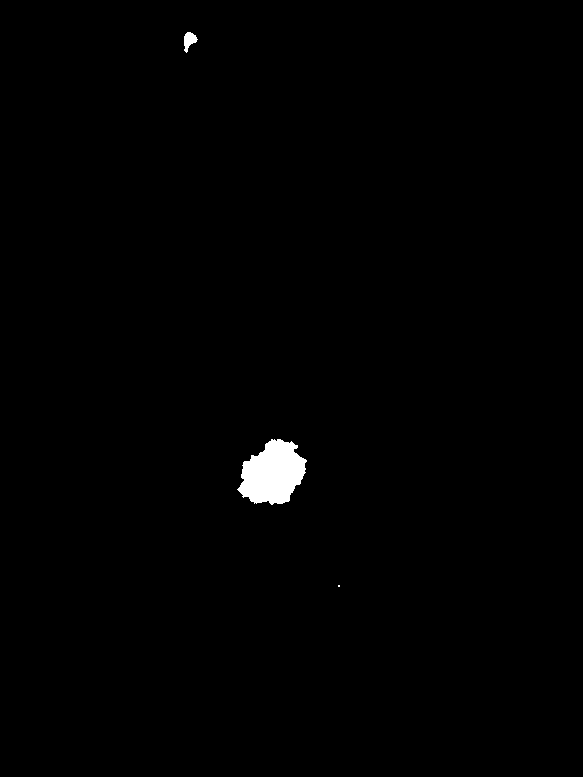}}
    \subfigure[]{\label{fig:6a}\includegraphics[width=0.1\textwidth]{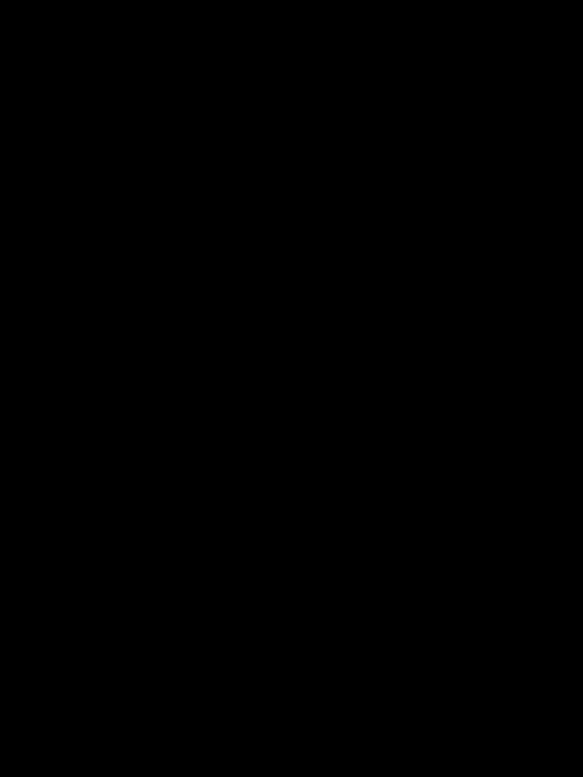}}
    \break
    \subfigure[]{\label{fig:7a}\includegraphics[width=0.1\textwidth]{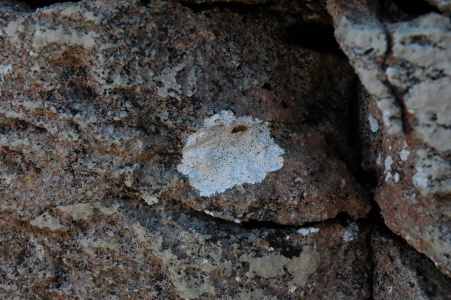}}
    \subfigure[]{\label{fig:8a}\includegraphics[width=0.1\textwidth]{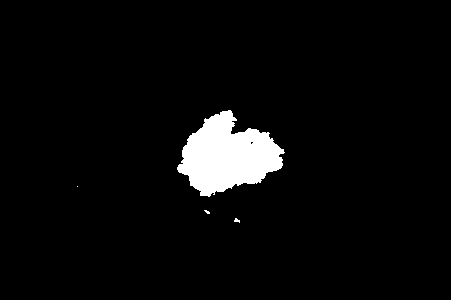}}
    \subfigure[]{\label{fig:9a}\includegraphics[width=0.1\textwidth]{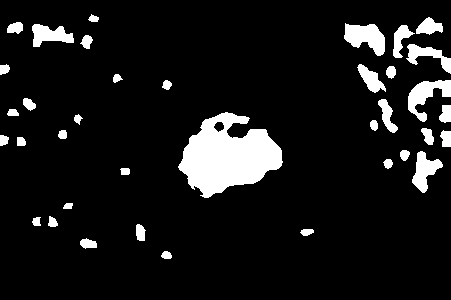}}
    \break
    \subfigure[]{\label{fig:10a}\includegraphics[width=0.1\textwidth]{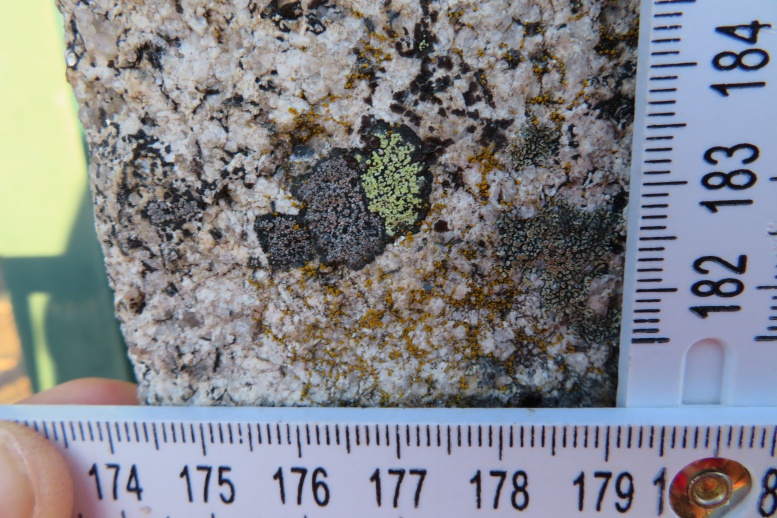}}
    \subfigure[]{\label{fig:11a}\includegraphics[width=0.1\textwidth]{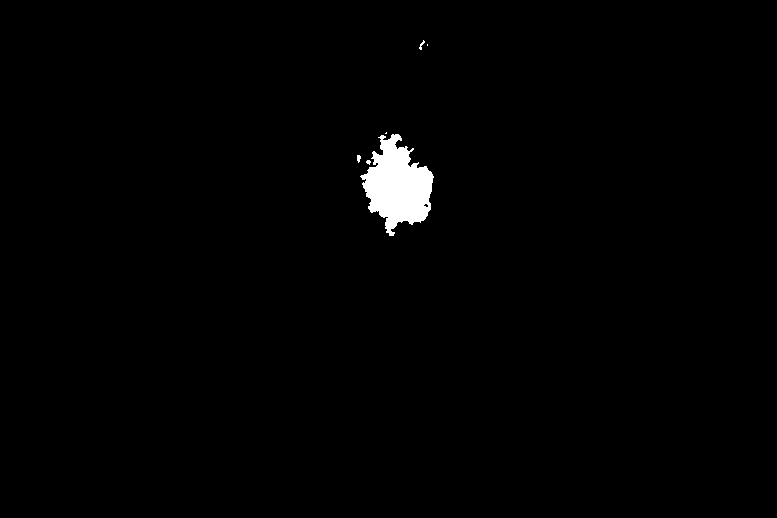}}
    \subfigure[]{\label{fig:12a}\includegraphics[width=0.1\textwidth]{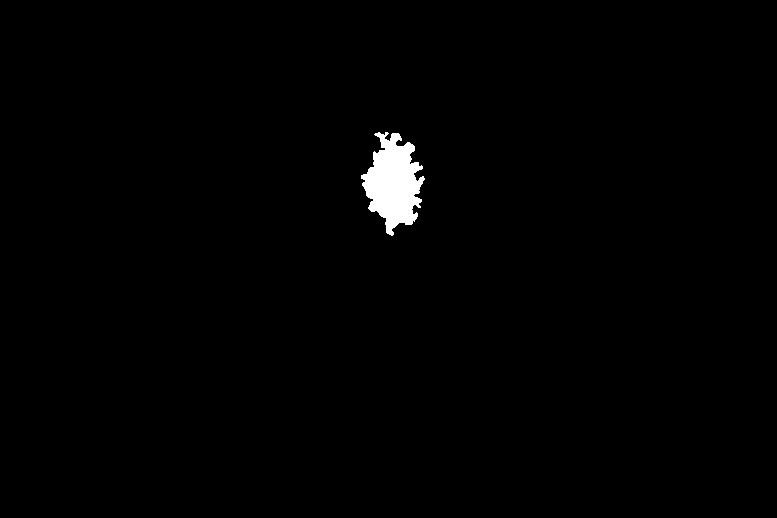}}
    \caption{Automatic classification for Antarctica (a, d), Cascais (g) and  Muro Escola Castelejo (j).}
    \label{fig:outputs}
\end{figure}

Following the previous section, some segmentation of the Antarctica, Cascais and Muro Escola Castelejo sets   (\Cref{fig:outputs} are presented.
The images are arranged so that, in each row, we first observe the original image, then the manual classification and finally an automatic classification. 
The automatic classifications of the  \Cref{fig:3a} and \Cref{fig:6a} (Antarctica) were produced using the {SVC} classifier, trained with 30 images and with the following {SLIC} parameters: threshold 0.5, n\_segments 500, compactness 20 and sigma 1. 

The automatic classification of  \Cref{fig:9a} (Cascais) was produced using  the RandomForestsClassifier classifier, trained with 27 images and with the following parameters: threshold 0.5, n\_segments 2000, compactness 10 and sigma 3. 

The automatic classification of  \Cref{fig:12a} (Muro Escola Castelejo) was produced using the {SVC} classifier, trained with 27 images and with the following parameters: threshold 0.5, n\_segments 500, compactness 20 and sigma 1.

For the first group of 3 pictures (Antarctica),  it turns out that a perfect segmentation is obtained. However, it is important to note that, for exactly the same conditions but different image (\Cref{fig:3a}), that same classifier produces a completely black image without the detection of the lichen. This is an example of a false negative.

For the Cascais set (\Cref{fig:5} and \Cref{fig:7}), it can be seen that, regardless of the classifier and the {SLIC} parameters, the segmentations are all, with rare exceptions, of poor quality. In fact, in  \Cref{fig:9a}, several false positives are observed.

In the case of  \Cref{fig:11a}, the classification was even better than the GrabCut manual classification.

\section{Conclusion}

A set of tools was produced  to assist the acquisition and processing of lichenometric photographic samples.
Even without considering the automatic classification, it was found that both the manual classification tool and the automated system of acquisitions with targets, improve and speed up the process. 
Estimates of lichen cover areas on rocks are automatically obtained for a larger number of data points with lower user time. 
This result makes more robust statistical analyses possible, and will contribute to studies that elaborate lichen growth models for age estimation using lichenometry.

With regard to manual classification, it was observed that large gains were obtained in classification times.
 These gains are due to the fact that the GrabCut algorithm considers both color and proximity between pixels to perform the segmentation. 
It was also observed that the quality of the classifications is on par with the classifications produced with the other methods.

The automatic classifier was developed to work for unknown future image sets sets, so ideal parameters must can not previously fixed. The classifier and feature generation parameters should be defined on a per data set case, by the user or using automatic parameter estimation.

The automatic classification does not performs optimally with data sets containing images with a wide scale range (that affect the SLIC algorithm), containing shades or where taken with various light intensity.
This can be solved with field guidelines for the user and with future additional components to estimate more suitable parameters.

For future studies, solutions based on non-visible spectrum bands or even visible multi-spectral images  can also be explored.

\bibliographystyle{unsrt}

\bibliography{./Bibliography.bib}

\begin{thebibliography}{10}

\bibitem{palacios2021}
David Palacios, Manuel Rodr{\'\i}guez-Mena, Jos{\'e}~M
  Fern{\'a}ndez-Fern{\'a}ndez, Irene Schimmelpfennig, Luis~M Tanarro,
  Jos{\'e}~J Zamorano, Nuria Andr{\'e}s, Jose {\'U}beda, {\TH}orsteinn
  S{\ae}mundsson, Skafti Brynj{\'o}lfsson, et~al.
\newblock Reversible glacial-periglacial transition in response to climate
  changes and paraglacial dynamics: A case study from
  h{\'e}{\dh}insdalsj{\"o}kull (northern iceland).
\newblock {\em Geomorphology}, 388:107787, 2021.

\bibitem{birkeland1982subdivision}
Peter~W Birkeland.
\newblock Subdivision of holocene glacial deposits, ben ohau range, new
  zealand, using relative-dating methods.
\newblock {\em Geological Society of America Bulletin}, 93(5):433--449, 1982.

\bibitem{carrara1975holocene}
PE~Carrara and ANDREWS JT.
\newblock Holocene glacial/periglacial record; northern san juan mountains,
  southwest. colorado.
\newblock 1975.

\bibitem{garibotti2009lichenometric}
Irene~Adriana Garibotti and Ricardo Villalba.
\newblock Lichenometric dating using rhizocarpon subgenus rhizocarpon in the
  patagonian andes, argentina.
\newblock {\em Quaternary research}, 71(3):271--283, 2009.

\bibitem{hansen2008application}
Eric~Steen Hansen.
\newblock The application of lichenometry in dating of glacier deposits.
\newblock {\em Geografisk Tidsskrift-Danish Journal of Geography},
  108(1):143--151, 2008.

\bibitem{o2003rhizocarpon}
Michael~A O'Neal and Katherine~R Schoenenberger.
\newblock A rhizocarpon geographicum growth curve for the cascade range of
  washington and northern oregon, usa.
\newblock {\em Quaternary Research}, 60(2):233--241, 2003.

\bibitem{Orwin_et_al_2008}
John Orwin, KRISTA MCKINZEY, MICHAEL STEPHENS, and Andrew Dugmore.
\newblock Identifying moraine surfaces with similar histories using lichen size
  distributions and the u2 statistic, southeast iceland.
\newblock {\em Geografiska Annaler: Series A, Physical Geography}, 90:151 --
  164, 06 2008.

\bibitem{pendleton2017using}
Simon~L Pendleton, Jason~P Briner, Darrell~S Kaufman, and Susan~R Zimmerman.
\newblock Using cosmogenic 10be exposure dating and lichenometry to constrain
  holocene glaciation in the central brooks range, alaska.
\newblock {\em Arctic, Antarctic, and Alpine Research}, 49(1):115--132, 2017.

\bibitem{proctor1983sizes}
MCF Proctor.
\newblock Sizes and growth-rates of thalli of the lichen rhizocarpon
  geographicum on the moraines of the glacier de valsorey, valais, switzerland.
\newblock {\em The Lichenologist}, 15(3):249--261, 1983.

\bibitem{roberts2010establishing}
Stephen~J Roberts, Dominic~A Hodgson, Samantha Shelley, Jessica Royles, Huw~J
  Griffiths, Tara~J Deen, and Michael~AS Thorne.
\newblock Establishing lichenometric ages for nineteenth-and twentieth-century
  glacier fluctuations on south georgia (south atlantic).
\newblock {\em Geografiska Annaler: Series A, Physical Geography},
  92(1):125--139, 2010.

\bibitem{roof2011indirect}
Steven Roof and Al~Werner.
\newblock Indirect growth curves remain the best choice for lichenometry:
  evidence from directly measured growth rates from svalbard.
\newblock {\em Arctic, antarctic, and alpine research}, 43(4):621--631, 2011.

\bibitem{rosenwinkel2015limits}
Swenja Rosenwinkel, Oliver Korup, Angela Landgraf, and Atyrgul Dzhumabaeva.
\newblock Limits to lichenometry.
\newblock {\em Quaternary Science Reviews}, 129:229--238, 2015.

\bibitem{trenbirth2010lichen}
Hazel~E Trenbirth and John~A Matthews.
\newblock Lichen growth rates on glacier forelands in southern norway:
  preliminary results from a 25-year monitoring programme.
\newblock {\em Geografiska Annaler: Series A, Physical Geography},
  92(1):19--39, 2010.

\bibitem{doi:10.1177/030913338500900202}
John~L. Innes.
\newblock Lichenometry.
\newblock {\em Progress in Physical Geography: Earth and Environment},
  9(2):187--254, 1985.

\bibitem{ivy2008surface}
Susan Ivy-Ochs and Florian Kober.
\newblock Surface exposure dating with cosmogenic nuclides.
\newblock {\em E\&G Quaternary Science Journal}, 57(1/2):179--209, 2008.

\bibitem{gliganic2019osl}
LA~Gliganic, MC~Meyer, R~Sohbati, M~Jain, and S~Barrett.
\newblock Osl surface exposure dating of a lithic quarry in tibet: Laboratory
  validation and application.
\newblock {\em Quaternary Geochronology}, 49:199--204, 2019.

\bibitem{article}
William Locke and {Et al.}
\newblock A manual for lichenometry. british gcomorphological research group.
\newblock {\em Technical Bulletin}, 26:1--47, 01 1979.

\bibitem{osborn2015lichenometric}
Gerald Osborn, Daniel McCarthy, Aline LaBrie, and Randall Burke.
\newblock Lichenometric dating: science or pseudo-science?
\newblock {\em Quaternary Research}, 83(1):1--12, 2015.

\bibitem{matthews2011}
John~A. Matthews and Hazel~E. Trenbirth.
\newblock Growth rate of a very large crustose lichen (rhizocarpon subgenus)
  and its implications for lichenometry.
\newblock {\em Geografiska Annaler: Series A, Physical Geography},
  93(1):27--39, 2011.

\bibitem{85f9b2036d5d48ac8a21b8e7c275886e}
{Richard A.} Armstrong.
\newblock {\em The influence of environmental factors on the growth of lichens
  in the field}, volume~1, pages 1--18.
\newblock Springer, Germany, 2015.

\bibitem{NicoleM.Henry:2011}
Nicole~M. Henry.
\newblock {Measurement of growth in the lichen Rhizocarpon geographicum using a
  new photographic technique}.
\newblock Master's thesis, Faculty of Mathematics and Science, Brock University
  St. Catharines, Ontario, April 2011.

\bibitem{TheresaA.Bukovics:2016}
Theresa~A. Bukovics.
\newblock {Photogrammetric Exploration of Demographic Change in Juvenile
  Rhizocarpon geographicum Thalli}.
\newblock Master's thesis, Faculty of Mathematics and Sciences, Brock
  University St. Catharines, Ontario, 2016.

\bibitem{MCCARTHY2021107736}
Daniel~P. McCarthy.
\newblock A simple test of lichenometric dating using bidecadal growth of
  rhizocarpon geographicum agg. and structure-from-motion photogrammetry.
\newblock {\em Geomorphology}, 385:107736, 2021.

\bibitem{maoliveira:2020}
{Maria}~A. {Oliveira} and {Et al.}
\newblock Estimating the age and mechanism of boulder transport related with
  extreme waves using lichenometry.
\newblock {\em Progress in Physical Geography}, 2020.

\bibitem{doi:10.1080/15230430.2001.12003411}
Daniel~P. McCarthy and Kamil Zaniewski.
\newblock Digital analysis of lichen cover: A technique for use in lichenometry
  and licnenology.
\newblock {\em Arctic, Antarctic, and Alpine Research}, 33(1):107--113, 2001.

\bibitem{hill_1981}
D.J. Hill.
\newblock The growth of lichens with special reference to the modelling of
  circular thalli.
\newblock {\em The Lichenologist}, 13(3):265–287, 1981.

\bibitem{Seminara}
Agnese Seminara and {Et al.}
\newblock A universal growth limit for circular lichens.
\newblock {\em Journal of The Royal Society Interface}, 15:20180063, 06 2018.

\bibitem{aghababaei2021classification}
Masoumeh Aghababaei, Ataollah Ebrahimi, Ali~Asghar Naghipour, Esmaeil Asadi,
  and Jochem Verrelst.
\newblock Classification of plant ecological units in heterogeneous semi-steppe
  rangelands: Performance assessment of four classification algorithms.
\newblock {\em Remote Sensing}, 13(17):3433, 2021.

\bibitem{hurskainen2019auxiliary}
Pekka Hurskainen, Hari Adhikari, Mika Siljander, PKE Pellikka, and Andreas
  Hemp.
\newblock Auxiliary datasets improve accuracy of object-based land use/land
  cover classification in heterogeneous savanna landscapes.
\newblock {\em Remote sensing of environment}, 233:111354, 2019.

\bibitem{akar2012classification}
{\"O}zlem Akar and Oguz G{\"u}ng{\"o}r.
\newblock Classification of multispectral images using random forest algorithm.
\newblock {\em Journal of Geodesy and Geoinformation}, 1(2):105--112, 2012.

\bibitem{wang2012machine}
Shijun Wang and Ronald~M Summers.
\newblock Machine learning and radiology.
\newblock {\em Medical image analysis}, 16(5):933--951, 2012.

\bibitem{gupta2018feature}
Vibha Gupta and Arnav Bhavsar.
\newblock Feature importance for human epithelial (hep-2) cell image
  classification.
\newblock {\em Journal of Imaging}, 4(3):46, 2018.

\bibitem{6910608}
Mürvet Kırcı and {Et al.}
\newblock Vegetation measurement using image processing methods.
\newblock In {\em 2014 The Third Intl. Conference on Agro-Geoinformatics},
  pages 1--5, 2014.

\bibitem{plant_diseases}
Jayme Barbedo.
\newblock Digital image processing techniques for detecting, quantifying and
  classifying plant diseases.
\newblock {\em SpringerPlus}, 2:660, 12 2013.

\bibitem{4310076}
N.~{Otsu}.
\newblock A threshold selection method from gray-level histograms.
\newblock {\em IEEE Transactions on Systems, Man, and Cybernetics},
  9(1):62--66, 1979.

\bibitem{LinKaiyan:2014}
Chen Jie Si~Huiping Lin~Kaiyan, Wu~JunHui.
\newblock Measurement of plant leaf area based on computer vision.
\newblock In {\em 2014 Sixth Intl. Conference on Measuring Technology and
  Mechatronics Automation}, 2014ffff.

\bibitem{salehi:2016}
Sara Salehi and {Et al.}
\newblock Identification of a robust lichen index for the deconvolution of
  lichen and rock mixtures using pattern search algorithm (case study:
  Greenland).
\newblock {\em ISPRS - Intl. Archives of the Photogrammetry, Remote Sensing and
  Spatial Information Sciences}, XLI-B7:973--979, 06 2016.

\bibitem{937505}
Y.Y. Boykov and M.-P. Jolly.
\newblock Interactive graph cuts for optimal boundary region segmentation of
  objects in n-d images.
\newblock In {\em Proceedings Eighth IEEE Intl. Conference on Computer Vision.
  ICCV 2001}, volume~1, pages 105--112 vol.1, 2001.

\bibitem{10.1145/1015706.1015720}
Carsten Rother and {Et al.}
\newblock "grabcut": Interactive foreground extraction using iterated graph
  cuts.
\newblock {\em ACM Trans. Graph.}, 23(3):309–314, August 2004.

\bibitem{boykov2001experimental}
Yuri Boykov and Vladimir Kolmogorov.
\newblock An experimental comparison of min-cut/max-flow algorithms for energy
  minimization in vision.
\newblock In {\em International workshop on energy minimization methods in
  computer vision and pattern recognition}, pages 359--374. Springer, 2001.

\bibitem{kim2008estimation}
M~Kim, M~Madden, and T~Warner.
\newblock Estimation of optimal image object size for the segmentation of
  forest stands with multispectral ikonos imagery.
\newblock In {\em Object-based image analysis}, pages 291--307. Springer, 2008.

\bibitem{slic_powaaa}
Radhakrishna Achanta and {Et al.}
\newblock Slic superpixels.
\newblock {\em Technical report, EPFL}, 06 2010.

\bibitem{hastie_09_elements-of.statistical-learning}
Trevor Hastie and {Et al.}
\newblock {\em The elements of statistical learning: data mining, inference and
  prediction}.
\newblock Springer, 2 edition, 2008.

\bibitem{scikit-learn}
{1.4 Support Vector Machines, scikit-learn 0.20.2 documentation}, 2020.

\bibitem{Boser92atraining}
Bernhard~E. Boser and {Et al.}
\newblock A training algorithm for optimal margin classifiers.
\newblock In {\em Proceedings of the 5th Annual ACM Workshop on Computational
  Learning Theory}, pages 144--152. ACM Press, 1992.

\bibitem{press2007numerical}
W.H. Press and {Et al.}
\newblock {\em Numerical Recipes 3rd Edition: The Art of Scientific Computing},
  volume Section 16.5. Support Vector Machines.
\newblock Cambridge University Press, 2007.

\bibitem{Breiman1996}
Leo Breiman.
\newblock Bagging predictors.
\newblock {\em Machine Learning}, 24(2):123--140, Aug 1996.

\bibitem{790410}
D.G. Lowe.
\newblock Object recognition from local scale-invariant features.
\newblock In {\em Proceedings of the Seventh IEEE Intl. Conference on Computer
  Vision}, volume~2, pages 1150--1157 vol.2, 1999.

\bibitem{MATTHEWS1975442}
B.W. Matthews.
\newblock Comparison of the predicted and observed secondary structure of t4
  phage lysozyme.
\newblock {\em Biochimica et Biophysica Acta (BBA) - Protein Structure},
  405(2):442--451, 1975.

\end{thebibliography}

\end{document}